\documentclass[preprint,12pt]{elsarticle}

\usepackage{graphicx}
\usepackage{amssymb}
\usepackage{multirow}
\usepackage{booktabs}
\usepackage{amssymb}
\usepackage{mathrsfs}
\usepackage{longtable}
\usepackage{lscape}
\usepackage{tabularx}
\usepackage{epsfig}
\usepackage{colortbl}
\usepackage{amsthm}
\usepackage{amsmath}
\usepackage{mathtools}
\newdefinition{definition}{Definition}
\newdefinition{example}{Example}
\newdefinition{remark}{Remark}
\newtheorem{theorem}{Theorem}[section]

\biboptions{square,sort&compress}

\journal{xx}

\begin{document}

\begin{frontmatter}



\title{A total uncertainty measure for D numbers based on belief intervals}

\author[NWPU]{Xinyang Deng\corref{COR}}
\ead{xinyang.deng@nwpu.edu.cn}
\author[NWPU]{Wen Jiang\corref{COR}}
\ead{jiangwen@nwpu.edu.cn}

\cortext[COR]{Corresponding author.}

\address[NWPU]{School of Electronics and Information, Northwestern Polytechnical University, Xi'an 710072, China}

\begin{abstract}
As a generalization of Dempster-Shafer theory, the theory of D numbers is a new theoretical framework for uncertainty reasoning. Measuring the uncertainty of knowledge or information represented by D numbers is an unsolved issue in that theory. In this paper, inspired by distance based uncertainty measures for Dempster-Shafer theory, a total uncertainty measure for a D number is proposed based on its belief intervals. The proposed total uncertainty measure can simultaneously capture the discord, and non-specificity, and non-exclusiveness involved in D numbers. And some basic properties of this total uncertainty measure, including range, monotonicity, generalized set consistency, are also presented.
\end{abstract}

\end{frontmatter}

\section{Introduction}
Uncertainty exists extensively in the real world, and many theories have been developed for representing and dealing with uncertainty, such as probability theory, imprecise probabilities, fuzzy logic, possibility theory, Dempster-Shafer theory, set-membership approach, and so on. For any theory of uncertainty reasoning, the quantification of uncertainty degree to a piece of information is a crucial issue \cite{klir1991generalized,klir2005uncertainty}. Regarding this issue, different branches of theories have given different solutions. For example, the Shannon entropy is widely accepted to be the uncertainty measure of probabilities. However, in many other uncertainty reasoning theories, there are not well acknowledged and satisfactory measures that can effectively quantify the uncertainty of information.

In this paper, we put our attention on a special theory for uncertainty reasoning which is called D numbers theory (DNT) \cite{deng2012DJICS99,deng2017d09186}. This theory is a recently proposed theoretical framework that has generalized Dempster-Shafer theory (DST) \cite{Dempster1967,Shafer1976}. In DST, every piece of knowledge or information is abstracted as a mass function or basic probability assignment (BPA), and each BPA is defined on a mutually exclusive and collectively exhaustive set which is called frame of discernment (FOD). In contrast, DNT generalizes the DST to a set with non-exclusive elements and does not strictly require the knowledge or information is complete. For more details about DNT and its differences with DST, please refer to references \cite{xydeng2017DNCR,deng2014environmental412,deng2017d09186} and introductions given in the next section. As for this new theory DNT, how to measure the uncertainty of D numbers has not been studied yet.

Since DNT is a generalization of DST, previous studies about uncertainty measures in DST would be useful for the design of uncertainty measures in DNT. In DST, measuring the uncertainty degree of a BPA is still an open issue. Some representative total uncertainty measures are aggregated uncertainty (AU) \cite{harmanec1994measuring224}, ambiguity measure (AM) \cite{jousselme2006measuring}, and other entropies discussed in \cite{abellan2008requirements,jirouvsek2018new,deng2016deng91}, and recently proposed belief interval based uncertainty measures \cite{yang2016new94,AnImprovedAPIN2017,wang2017uncertaintySY}, to name but a few. These uncertainty measures have studied the uncertainty quantification of BPAs from different perspectives for example axiomatization, desirable properties and behaviour, consistency with probabilities or only taking into account the framework of DST. These perspectives and existing uncertainty measures for DST are also valuable for the design of an uncertainty measure for D numbers. However, different from DST that only contains two types of uncertainty factors which are discord and non-specificity \cite{yager1983entropy94249}, in DNT there exists a new type of uncertainty caused by the non-exclusiveness among elements. In addition, since a D number can be information-incomplete which means that the knowledge or information is incomplete, a rational uncertainty measure for D numbers must be able to deal with the case of incomplete information or knowledge.

In this paper, inspired by distance-based uncertainty measures \cite{yang2016new94,AnImprovedAPIN2017} for BPAs in DST, a total uncertainty measure for D numbers, denoted as $TU$, is proposed based on the belief intervals of D numbers. The proposed $TU$ can simultaneously capture the discord, and non-specificity, and non-exclusiveness involved in a D number. Meanwhile, the possible incompleteness of information or knowledge is also considered in $TU$ by introducing a new notation $X$ representing unknown event. Moreover, some basic properties of $TU$ including range, monotonicity, generalized set consistency, are presented.

The rest of this paper is organized as follows. Section \ref{SectPreliminaries} gives a brief introduction about DST and DNT. In Section \ref{SectProposedTU}, a total uncertainty measure for D numbers $TU$ is proposed and basic properties of $TU$ is presented. Finally, Section \ref{SectConclusion} concludes the paper.

\section{Preliminaries}\label{SectPreliminaries}
\subsection{Basics of Dempster-Shafer theory}
Dempster-Shafer theory (DST) \cite{Dempster1967,Shafer1976}, also called belief function theory or evidence theory, is a popular tool for uncertainty reasoning because of its advantages in expressing uncertainty. As a theory of reasoning under the uncertain environment, DST has an advantage of directly expressing the ``uncertainty" by assigning the basic probability to a set composed of multiple objects, rather than to each of the individual objects. For completeness of the explanation, a few basic concepts in DST are introduced as follows.

Let $\Omega$ be a set of $N$ mutually exclusive and collectively exhaustive events, indicated by
\begin{equation}
\Omega  = \{ q_1 ,q_2 , \cdots ,q_i , \cdots ,q_N \}
\end{equation}
where set $\Omega$ is called a frame of discernment (FOD). The power set of $\Omega$ is indicated by $2^\Omega$, namely
\begin{equation}
2^\Omega   = \{ \emptyset ,\{ q_1 \} , \cdots ,\{ q_N \} ,\{ q_1
,q_2 \} , \cdots ,\{ q_1 ,q_2 , \cdots ,q_i \} , \cdots ,\Omega \}.
\end{equation}
The elements of $2^\Omega$ or subsets of $\Omega$ are called propositions.

\begin{definition}
Let a FOD be $\Omega = \{ q_1 ,q_2 , \cdots, q_N \}$, a mass function defined on $\Omega$ is a mapping $m$ from  $2^\Omega$ to $[0,1]$, formally defined by:
\begin{equation}
m: \quad 2^\Omega \to [0,1]
\end{equation}
which satisfies the following condition:
\begin{eqnarray}
m(\emptyset ) = 0 \quad {\rm{and}} \quad \sum\limits_{A \subseteq \Omega }{m(A) = 1}.
\end{eqnarray}
\end{definition}
In DST, a mass function is also called a basic probability assignment (BPA). The assigned basic probability $m(A)$ measures the belief exactly assigned to $A$ and represents how strongly the evidence supports $A$. If $m(A) > 0$, $A$ is called a focal element, and the union of all focal elements is called the core of the mass function.

Given a BPA, its associated belief measure $Bel_{m}$ and plausibility measure $Pl_{m}$ express the lower bound and upper bound of the support degree to each proposition in that BPA, respectively. They are defined as
\begin{equation}
Bel_{m} (A) = \sum\limits_{B \subseteq A} {m(B)},
\end{equation}
\begin{equation}
Pl_{m}(A) = \sum\limits_{B \cap A \ne \emptyset }{m(B)},
\end{equation}
Obviously, $Pl_{m}(A) \ge Bel_{m}(A)$ for each $A \subseteq \Omega$, and $[Bel_{m}(A), Pl_{m}(A)]$ is called the belief interval of $A$ in $m$.

\subsection{D numbers theory}
D numbers theory (DNT) is a new theoretical framework for uncertainty reasoning which generalizes the DST from two aspects: on one hand, the elements within FOD are not required to be mutually exclusive in DNT; on the other hand, the providing information in DNT can be incomplete in contrast to $\sum {m( \cdot )}  = 1$ in DST. For more theoretical details about DNT and its recent advances, please refer to literatures \cite{deng2012DJICS99,xydeng2017DNCR,deng2017d09186}. Furthermore, some applications of DNT can be found in references \cite{fan2016hybrid44,liu2014failure4110,xiao2016intelligent3713518,wang2017modifiedIJFS,deng2014environmental412,Deng2017Fuzzy2086}. A few of basic concepts in DNT are introduced as follows \cite{deng2017d09186}.

\begin{definition}\label{DefDNumbers}
Let $\Theta$ be a nonempty finite set $\Theta  = \{ \theta_1 ,\theta_2 , \cdots ,\theta_N \}$, a D number is a mapping formulated by
\begin{equation}
D: 2^{\Theta} \to [0,1]
\end{equation}
with
\begin{eqnarray}
\sum\limits_{B \subseteq \Theta } {D(B)} \le 1  \quad {\rm {and}} \quad
D(\emptyset ) = 0
\end{eqnarray}
where $\emptyset$ is the empty set and $B$ is a subset of $\Theta$.
\end{definition}

It is worthy noting that a D number can be defined on a set with non-exclusive elements, which means that any pair of elements in $\Theta$, for example $\theta_i, \theta_j \in \Theta$, are not required to be strictly exclusive, i.e. $\theta_i  \cap \theta_j  \ne \emptyset$. Here, we still call $\Theta$ as a FOD, but should note that a FOD in DNT is a set consisting of non-exclusive elements. Besides, according to Definition \ref{DefDNumbers}, in a D number the information is not required to be complete. If $\sum\limits_{B \subseteq \Theta } {D(B) = 1}$, we say that the D number is information-complete. By contrast, if $\sum\limits_{B \subseteq \Theta } {D(B) < 1}$ the D number is information-incomplete. In order to transform a D number with incomplete information to the information-complete case, a new notation ${ X}$ is imported to represent the unknown event, and there is not any restriction for the relationship between $X$ and $\Theta$. So, a new definition about D numbers is given as below.
\begin{definition}\label{DefDNumbersEquivalent}
A D number defined on a nonempty finite set $\Theta  = \{ \theta_1 ,\theta_2 , \cdots ,\theta_N \}$ and unknown event $X$ is a mapping $D: 2^{\Theta \cup X} \to [0,1]$ satisfying
\begin{eqnarray}
{\sum\limits_{B \subseteq {\Theta \cup X} } {D(B)} } = 1  \quad {\rm {and}} \quad
D(\emptyset ) = 0
\end{eqnarray}
\end{definition}

In the latter definition of D numbers, i.e. Definition \ref{DefDNumbersEquivalent}, $X$ represents the unknown (or incomplete) part in a D number. Regarding the requirement of non-exclusiveness in DNT, a membership function is developed to measure the non-exclusive degrees in $\Theta  \cup X$.

\begin{definition}\label{DefNonExclusiveness}
Given $B_i ,B_j  \in 2^{\Theta  \cup X}$, the non-exclusive degree between $B_i$ and $B_j$ is characterized by a mapping $u_{\neg E}$:
\begin{equation}
u_{\neg E} :2^{\Theta  \cup X}   \times 2^{\Theta  \cup X}   \to [0,1]
\end{equation}
with
\begin{equation}
u_{\neg E} (B_i ,B_j ) = \left\{ \begin{array}{l}
 1,\quad B_i  \cap B_j  \ne \emptyset   \\
 p,\quad B_i  \cap B_j  = \emptyset   \\
 \end{array} \right.
\end{equation}
and
\begin{equation}
u_{\neg E} (B_i ,B_j ) = u_{\neg E} (B_j ,B_i )
\end{equation}
where $0 \le p \le 1$. If letting the exclusive degree between $B_i$ and $B_j$ be denoted as $u_{E}$, then $u_{E} = 1 - u_{\neg E}$.
\end{definition}

According to Definition \ref{DefNonExclusiveness}, the non-exclusive degree between $B_i$ and $B_j$ is 1 if $B_i$ and $B_j$ have intersections, otherwise $u_{\neg E} (B_i ,B_j )$ is $p$ taking a value from $[0,1]$. Obviously, if $u_{\neg E} (B_i ,B_j ) = 0$ for any $B_i  \cap B_j  = \emptyset$, the FOD $\Theta$ in DNT is degenerated to classical FOD in DST.

In order to express the bound of uncertainty in a D number, in a very recent study \cite{deng2017d09186} we have developed a belief measure and a plausibility measure for D numbers.

\begin{definition}
Let $D$ represent a D number defined on $\Theta \cup X$ where $X$ represents unknown event, for any proposition $A \subseteq \Theta \cup X$, its belief measure $Bel: 2^{\Theta  \cup X}   \to [0,1]$ is defined as
\begin{equation}
Bel(A) = \sum\limits_{B \subseteq A} {D(B)},
\end{equation}
and its plausibility measure $Pl: 2^{\Theta  \cup X}  \to [0,1]$ is defined as
\begin{equation}
Pl(A) = \sum\limits_{B \cap A \ne \emptyset } {D(B)}  + \sum\limits_{B \cap A = \emptyset } {u_{\neg E} (B,A)D(B)},
\end{equation}
where $B \subseteq \Theta \cup X$.
\end{definition}

For the above definition, because $u_{\neg E} (B ,A ) = 1$ for $B  \cap A  \ne \emptyset$, the plausibility measure $Pl$ can also be written as
\begin{equation}
Pl(A) = \sum\limits_{B \subseteq \Theta  \cup X} {u_{\neg E} (B,A)D(B)}.
\end{equation}
As same as DST, $[Bel(A), Pl(A)]$ is called the belief interval of $A$ in DNT, which expresses the lower bound and upper bound of support degree to proposition $A$. And it is easy to find that the $Bel$ and $Pl$ for D numbers will degenerate to classical belief measure and plausibility measure in DST if the associated D number is a BPA in fact.

\section{Proposed total uncertainty measure for D numbers}\label{SectProposedTU}
\begin{figure}[htbp]
\begin{center}
\psfig{file=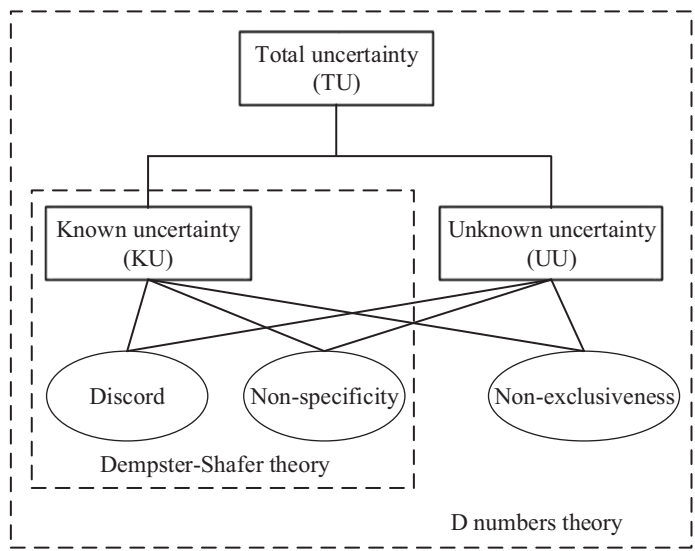,scale=0.5}
\caption{Uncertainty in D numbers}\label{FigUncertainties}
\end{center}
\end{figure}

How to measure the uncertainty of information is an important issue in the theories of uncertainty reasoning. Up to now, uncertainty measure for D numbers is an unsolved problem in DNT. With respect to this problem, there are two considerable aspects, as graphically shown in Figure \ref{FigUncertainties}. At first, since a D number consists of two parts, $\Theta$ called known part and $X$ called unknown part, a rational uncertainty measure must be able to model the total uncertainty that contains known uncertainty caused by known part and unknown uncertainty from unknown part. This is the first difference between DNT and DST in the design of uncertainty measure. At second, in DST the uncertainty consists of discord and non-specificity \cite{yager1983entropy94249}, in DNT, however, the non-exclusiveness among elements in ${\Theta  \cup X}$ becomes a new source of uncertainty. Therefore, a rational uncertainty measure for D numbers should simultaneously capture discord, non-specificity, and non-exclusiveness.

In this paper, a belief interval based total uncertainty measure, called $TU$, is proposed for D numbers. This idea of $TU$ is inspired by distance-based uncertainty measures \cite{yang2016new94,AnImprovedAPIN2017} for mass functions in DST.

\begin{definition}
Let $D$ represent a D number defined on $\Theta \cup X$ where $ \Theta = \{ \theta_1 ,\theta_2 , \cdots ,\theta_N \}$ and $X$ expresses the unknown part in $D$, the total uncertainty of $D$ is
\begin{equation}\label{EqTU}
TU(D) = KU(D) + UU(D)
\end{equation}
with
\begin{equation}
{KU(D)} = \sum\limits_{\theta_i \in \Theta} {\left[ {1 - d_E^I \left( {[Bel(\theta _i ),Pl(\theta _i )],[0,1]} \right)} \right]}
\end{equation}
\begin{equation}
UU(D) = Pl(X) \cdot U(X)
\end{equation}
where
\begin{equation}
d_E^I \left( {[Bel(\theta _i ),Pl(\theta _i )],[0,1]}\right) = \sqrt {\left[ {Bel(\theta _i )  - 0 } \right]^2  + \left[ {Pl(\theta _i )  - 1 } \right]^2 }
\end{equation}
and $U(X)$ is a function associated with the cardinality of $X$ and expresses the overall uncertainty in $X$, $Bel$ and $Pl$ are the belief measure and plausibility measure of D numbers, respectively.
\end{definition}

The underlying assumption of this total uncertainty measure is that an element in $\Theta \cup X$ has the largest uncertainty degree if its belief interval is $[0,1]$. Let us analyze the $TU$:

\begin{itemize}
  \item It separately calculates the known uncertainty and unknown uncertainty involved in a D number.
  \item For the known uncertainty, $KU$ employs the Euclidean distance function $d_E^I$ to calculate the distance between belief interval of each singleton $\theta_i$ (i.e., $[Bel(\theta _i ),Pl(\theta _i )]$) and the most uncertain interval $[0,1]$, then expresses the contribution of $\theta_i$ to the total uncertainty of $D$ by using ${1 - d_E^I \left( {[Bel(\theta _i ),Pl(\theta _i )],[0,1]} \right)}$.
  \item For the unknown uncertainty, a fictitious function $U(X)$ is assumed to represent the overall uncertainty of $X$, and $U(X)$ is only associated the the size of $X$. Since $X$ is an unknown set, we suppose $|X| \ge 2$ and $u_{\neg E} (x_i ,B) = u_{\neg E} (X,B)$ for any $x_i \in X$ and $B \subseteq \Theta \cup X$. Then, the belief interval of $x_i$ is calculated to be $[0, Pl(X)]$, so in $D$ the contribution of $x_i$ to the total uncertainty is $Pl(X)$ and the overall unknown uncertainty is positively correlated with $Pl(X) \cdot |X|$, namely $UU(D) = Pl(X) \cdot U(X)$. The principle behind $UU$ is logically consistent with that of $KU$.
  \item Discord, and non-specificity, and non-exclusiveness, are all considered in constructing belief intervals of singletons and simultaneously captured in $TU$.
\end{itemize}

\begin{figure}[htbp]
\begin{center}
\psfig{file=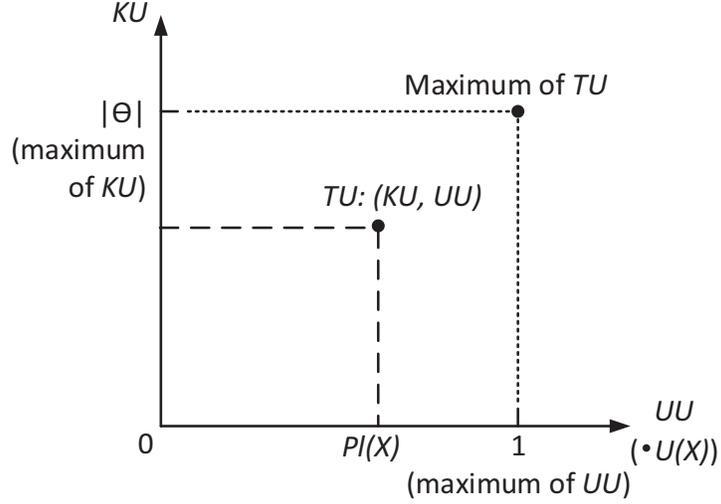,scale=0.65}
\caption{Graphical representation of $TU$}\label{FigTUgraph}
\end{center}
\end{figure}

Based on the above analysis, for the sake of simplicity the total uncertainty of a D number, $TU$, can be represented by a tuple $(KU, UU)$, as graphically shown in Figure \ref{FigTUgraph}, where $UU$ is a coefficient with respect to $U(X)$. Some basic properties of the proposed $TU$ are given as follows.

\begin{theorem}
\textbf{Range.} Given a D number $D$ defined over $\Theta \cup X$, the total uncertainty of $D$, denoted as $(KU(D), UU(D))$, is limited by ranges $0 \le KU(D) \le |\Theta|$ and $0 \le UU(D) \le 1$.
\end{theorem}

\begin{theorem}
\textbf{Monotonicity.} Let $D_1$ and $D_2$ be two arbitrary D numbers defined on $\Theta \cup X$, if
\[
\forall A \subseteq \Theta  \cup X: \quad \left[ {Bel_{D_1 } (A),Pl_{D_1 } (A)} \right] \subseteq \left[ {Bel_{D_2 } (A),Pl_{D_2 } (A)} \right]
\]
then
\[
KU(D_1 ) \le KU(D_2 )\quad {\rm{and}}\quad UU(D_1 ) \le UU(D_2 )
\]
\end{theorem}

\begin{theorem}
\textbf{Generalized set consistency.} When a set $A$, $A \subseteq \Theta$ exists such that $D(A) = 1$ then
\[KU(D) = |A| + \sum\limits_{{\theta _i} \in \Theta \backslash A} {{u_{\neg E}}({\theta _i},A)} \]
namely
\[KU(D) \propto \;|A|\]
where $\Theta \backslash A$ is the difference between $\Theta$ and $A$.
\end{theorem}

\section{Conclusion}\label{SectConclusion}
This paper has studied the issue of uncertainty quantification of knowledge or information in DNT. At first, three types of uncertainty factors, namely discord, and non-specificity, and non-exclusiveness, are identified for D numbers. Then, on the basis of our previous defined belief intervals for D numbers and distance based uncertainty measures for DST, a total uncertainty measure $TU$ is proposed in the framework of D numbers theory. At last, some basic properties of $TU$ are presented. In the future study, on one hand other properties of $TU$ will be investigated, on the other hand the applications of the proposed $TU$ are given more attention.

\section*{Acknowledgments}
The work is partially supported by National Natural Science Foundation of China (Program Nos. 61703338, 61671384), Natural Science Basic Research Plan in Shaanxi Province of China (Program No. 2016JM6018), Project of Science and Technology Foundation, Fundamental Research Funds for the Central Universities (Program No. 3102017OQD020).





\bibliographystyle{elsarticle-num}
\bibliography{references}







\end{document}